\documentclass[10pt,twocolumn,letterpaper]{article}

\usepackage{wacv}              

\usepackage[accsupp]{axessibility}
\usepackage{graphicx}
\usepackage{amsmath}
\usepackage{amssymb}
\usepackage{booktabs}

\usepackage[pagebackref,breaklinks,colorlinks]{hyperref}

\usepackage[capitalize]{cleveref}
\crefname{section}{Sec.}{Secs.}
\Crefname{section}{Section}{Sections}
\Crefname{table}{Table}{Tables}
\crefname{table}{Tab.}{Tabs.}

\usepackage{todonotes}
\usepackage{xcolor}
\usepackage{graphicx}

\usepackage{rotating}
\usepackage{tabularx}
\usepackage{amssymb}
\usepackage{pifont}
\usepackage{adjustbox}
\newcommand{\xmark}{\ding{55}}%
\usepackage{booktabs}

\newcommand{\greencheck}{{\color{green}\checkmark}}
\newcommand{\redxmark}{{\color{red}\xmark}}

\usepackage[acronym]{glossaries}
\newacronym{ALS}{ALS}{airborne laser scanning}
\newacronym{MLS}{MLS}{mobile laser scanning}
\newacronym{LoD}{LoD}{level of detail}
\newacronym{LoDs}{LoDs}{level of details}
\newacronym{OGC}{OGC}{Open Geospatial Consortium}
\newacronym{GML}{GML}{Geography Markup Language}
\newacronym{ASAM}{ASAM}{Association for Standardization of Automation and Measuring Systems}
\newacronym{TLS}{TLS}{terrestrial laser scanning}
\newacronym{UAV}{UAV}{unmanned aerial vehicle}
\newacronym{HD}{HD}{high definition}
\newacronym{RANSAC}{RANSAC}{RANdom SAmple Consensus}
\newacronym{ROI}{ROI}{region of interest}
\newacronym{DEM}{DEM}{digital elevation model}
\newacronym{ICP}{ICP}{iterative closest point}
\newacronym{NLOS}{NLOS}{non-line-of-sight}
\newacronym{SfM}{SfM}{structure from motion}
\newacronym{FME}{FME}{Feature Manipulation Engine}
\newacronym{OSM}{OSM}{OpenStreetMap} 
\newacronym{RMSE}{RMSE}{root mean square error}
\newacronym{CPT}{CPT}{conditional probability table}
\newacronym{DST}{DST}{Dempster–Shafer theory}
\newacronym{BN}{BayNet}{Bayesian network}
\newacronym{GIS}{GIS}{Geographic Information System}
\newacronym{PPD}{PPD}{posterior probability distribution}
\newacronym{CI}{CI}{confidence interval}
\newacronym{IFC}{IFC}{Industry Foundation Classes}
\newacronym{CRS}{CRS}{coordinate reference system}
\newacronym{LoFG}{LoFG}{Level of Facade Generalization}

\def\wacvPaperID{1834} 
\def\confName{WACV}
\def\confYear{2025}

\begin{document}

\title{ZAHA: Introducing the Level of Facade Generalization and the Large-Scale Point Cloud Facade Semantic Segmentation Benchmark Dataset}


\author{Olaf Wysocki\textsuperscript{1 }, Yue Tan\textsuperscript{1 }, Thomas Froech\textsuperscript{1 }, Yan Xia\textsuperscript{1,3 },  Magdalena Wysocki\textsuperscript{1,3 }, \\ Ludwig Hoegner\textsuperscript{2 }, Daniel Cremers\textsuperscript{1,3 }, Christoph Holst\textsuperscript{1 } \\ \\ \textsuperscript{1 }Technical University of Munich, \textsuperscript{2 }Munich University of Applied Sciences, \\ \textsuperscript{3 }Munich Center for Machine Learning (MCML)\\
\tt\small (olaf.wysocki, yue.tan, thomas.froech, yan.xia, magdalena.wysocki, \\ 
\tt\small ludwig.hoegner, cremers, christoph.holst)@tum.de }



\maketitle

\begin{abstract}
Facade semantic segmentation is a long-standing challenge in photogrammetry and computer vision. 
Although the last decades have witnessed the influx of facade segmentation methods, there is a lack of comprehensive facade classes and data covering the architectural variability.
In ZAHA\footnote{Project page: https://github.com/OloOcki/zaha}, we introduce Level of Facade Generalization (LoFG), novel hierarchical facade classes designed based on international urban modeling standards, ensuring compatibility with real-world challenging classes and uniform methods' comparison.
Realizing the LoFG, we present to date the largest semantic 3D facade segmentation dataset, providing 601 million annotated points at five and 15 classes of LoFG2 and LoFG3, respectively. 
Moreover, we analyze the performance of baseline semantic segmentation methods on our introduced LoFG classes and data, complementing it with a discussion on the unresolved challenges for facade segmentation. 
We firmly believe that ZAHA shall facilitate further development of 3D facade semantic segmentation methods, enabling robust segmentation indispensable in creating urban digital twins.
\end{abstract}
%
%
\section{Introduction}
\label{sec:intro}
\sloppy
\begin{figure*}
    \centering
    \includegraphics[width=0.9\linewidth]{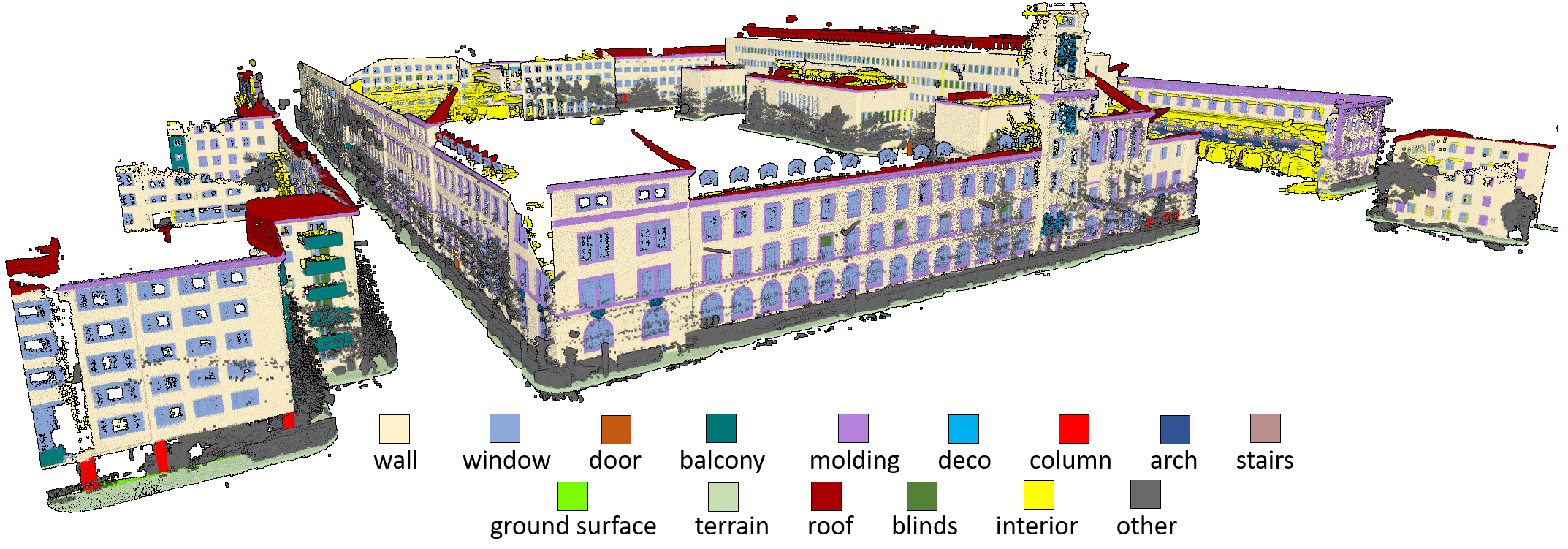}
    \caption{ZAHA: The dataset comprising 66 facades of various architectural style yielding over 601 million facade-level annotated terrestrial point clouds with distinct 15 facade-relevant classes.}
    \label{fig:catchy}
\end{figure*}
Facade semantic segmentation is a fundamental issue in photogrammetry and computer vision \cite{szeliski2010computer}.
The issue has became compounded thanks to such innovative architects as Zaha Hadid, challenging the standard assumptions of the wall's planarity, utilizing new materials, and mingling old with new architecture.

Throughout the years, various methods have been created for image-based facade segmentation, predominantly facilitated by datasets of annotated facade images \cite{musialski2013survey,liu2020deepfacade,korc2009etrims}.
Although the segmentation performance on the ortho-rectified images can reach even 90\% \cite{liu2020deepfacade}, 
the 2D-image nature hinders the immediate understanding of 3D scenes, limiting the 3D facade segmentation capabilities.
The striking example are intruded and extruded facade elements, which have been under-explored, owing to inability to capture their structure and depth with 2D ortho-rectified images.
These flaws impact other research fields frequently relying on a 3D facade semantic segmentation, such as 3D semantic building reconstruction at a high level of detail \cite{wysocki2023scan2lod3}, finding its applications in simulating flood risk~\cite{amirebrahimi2016bim} or testing automated driving functions~\cite{schwabRequirementAnalysis3d2019}, among others \cite{wysockiMLS2LoD3,willenborgIntegration2018}.

Recent developments have demonstrated that \gls*{MLS} devices can fill this 3D data gap: Once they are mounted on a mobile platform, they deliver dense, street-level point clouds, enabling capturing a 3D urban environment and, thus, 3D facade geometry (\cref{fig:catchy}) \cite{xu2021towards}.
This trait has sparked significant growth in urban point cloud benchmark datasets \cite{tumfacadePaper,klimkowska2022detailed} and the development of many semantic segmentation methods \cite{li2020deep,xu2021towards,tan2020toronto}.

Yet, one of the crucial impediments in developing facade segmentation methods is the facade classes heterogeneity; leading to misinterpretations of facade element characteristics and hampering the development of a large-scale, high-variability data.
This overlooked issue also impedes applying semantically segmented facades directly to the 3D semantic building reconstruction tasks since they mismatch taxonomically and geometrically the well-established modeling classes defined in the international standards \cite{wysocki2023scan2lod3,grogerOGCCityGeography2012,xu2021towards}.
Another key issue pertains to the limited annotated facade samples for methods' training and validation.
Only two relatively small datasets comprise facade-level classes:
TUM-FAÇADE \cite{tumfacadePaper} and ArCH \cite{archDatasetPaper}, which however either focus on a single university dataset or on only specific cultural heritage buildings, respectively.
To tackle the challenges mentioned above, in this paper, we present the following contributions:
%
\begin{itemize}
    \item Proposing \gls*{LoFG}: Novel hierarchical facade classes designed based on international facade-related standards;
    \item Realizing LoFG concept on the introduced, to date, the largest point cloud facade semantic segmentation dataset, boasting 601 million points (\cref{fig:catchy});
    \item Identifying outstanding challenges of the 3D facade semantic segmentation based on the extensive experimental and research analysis.
\end{itemize}
\section{Related Works}
\label{sec:related-work}
\subsection{Facade-Related Point Cloud Datasets}
For many years researchers have invested a great deal of effort in introducing multiple image-based facade segmentation datasets for methods development \cite{szeliski2010computer,riemenschneider2012irregular, korc2009etrims,Tylecek13,gadde2016learning}. 
For instance, in one the first works over a decade ago, the eTRIMS dataset has introduced 60 facade images~\cite{korc2009etrims} and eight facade-relevant labels.
Yet, these and following works \cite{Tylecek13} have been inconsistent with their facade elements definitions and have been limited concerning intruded and extruded 3D facade elements (e.g., arches).

\begin{table*}[!htb]
    \captionsetup{size=footnotesize}
    \caption{Point cloud benchmark datasets for testing urban semantic segmentation.} 
    \label{tab:bigTable}
    \centering
    \begin{adjustbox}{width=1\textwidth, center}
        \begin{tabular}{lccccccccc}
            \toprule
            Name  & Year &   Sensor   &   World   &  \# Classes   & Facade-level &  \# Facade-labeled & Various facade & Hierarchical \\
            &     &      &      &     & classes? & points & types? & classes?\\
            \midrule
            Oakland 3D \cite{Munoz-2009-10227}   &   2009   &   MLS   &   real   &   44   & $\thicksim$   &   $\thicksim$1.6 M  &  \redxmark  & \redxmark \\
            ETH PRS \cite{ethprs}  &   2012   &   TLS   &   real   &   0   & \redxmark &   \redxmark  & - & \redxmark  \\
            Sydney Urban Objects Dataset \cite{SydneyDatasetde2013unsupervised}  &   2013   &   MLS   &   real   &   26   & \redxmark &  \redxmark  & - & \redxmark \\
            Paris-rue-Madame database \cite{serna2014parisMadame}   &   2014   &   MLS   &   real   &   27   & $\thicksim$   &   $\thicksim$20 M  & \redxmark & \redxmark   \\
            iQumulus  \cite{vallet2015terramobilita} &   2015   &   MLS   &   real   &   8   & \redxmark &   \redxmark  & - & \redxmark  \\
            TUM-MLS-2016 \cite{zhu_tum-mls-2016_2020}  &   2016   &   MLS   &   real   &   9   & \redxmark &   \redxmark  & - & \redxmark  \\
            semantic3D.net \cite{hackel2017semantic3d}  &   2017   &   TLS   &   real   &   9   & \redxmark &   \redxmark  & - & \redxmark  \\
            Paris-Lille-3D \cite{roynard2018parisLille}  &   2018   &   MLS   &   real   &   50   & \redxmark &  \redxmark  & - & $\thicksim$  \\
            SynthCity \cite{griffiths2019synthcity}   &   2019   &   MLS   &   synthetic   &   9   & \redxmark &   \redxmark  & - & \redxmark  \\
            A2D2 \cite{geyer2020a2d2}  &   2020   &   MLS   &   real   &   38   & \redxmark &   \redxmark  & - & \redxmark  \\
            ArCH \cite{matrone2020comparing}  &   2020   &   TLS/MLS/UAV   &   real   &   10   & \greencheck &   136 M & \redxmark & \redxmark   \\
            Toronto-3D \cite{tan2020toronto}  &   2020   &   MLS   &   real   &   8   & \redxmark &   \redxmark & - & \redxmark  \\
            Whu-TLS \cite{dong2020registration}  &   2020   &   TLS   &   real   &   0   & \redxmark &   \redxmark  & -  & \redxmark \\
            BIMAGE Datasets \cite{BimageBlaser}  &   2021   &   MLS   &   real   &   0   & \redxmark &   \redxmark  & - & \redxmark \\
            KITTI-360 \cite{liao2021kitti}  &   2021   &   MLS   &   real   &   19   & \redxmark &   \redxmark  & -  & \redxmark \\
            Paris-CARLA-3D \cite{deschaud2021pariscarla3d}  &   2021   &   MLS   &   real/synthetic   &   23   & \redxmark &   \redxmark  & - & \redxmark \\
            TUM-FAÇADE \cite{tumfacadePaper}  &   2022   &   MLS   &   real   &   17   & \greencheck &   118 M & \redxmark & \redxmark  \\
            HelixNet \cite{helixnet}  &   2022   &   MLS   &   real   &   9   & \redxmark &   \redxmark  & - & \redxmark \\
            SUD \cite{SUDdata_SiliviaGonzalez}  &   2023   &   MLS   &   real   &   8   & \redxmark &   \redxmark  & - & \redxmark \\ \hline
            \textbf{ZAHA (ours)} &   2024   &   MLS   &   real   &   15   & \greencheck &   601 M  & \greencheck & \greencheck \\
            \bottomrule
        \end{tabular}
    \end{adjustbox}
\end{table*}

Unlike the image-based data, the point cloud datasets for facade segmentation remain in their infancy (see \Cref{tab:bigTable}).
Although many benchmarks are devoted to 3D urban semantic segmentation, only ArCH \cite{archDatasetPaper} and TUM-FAÇADE \cite{tumfacadePaper} datasets comprise facade-level classes.
Yet, the ArCH dataset focuses on cultural heritage buildings, which trait renders it infeasible for methods' testing on common facade types;
Whereas the TUM-FAÇADE represents solely university buildings, and its relatively small size of 14 facades makes it insufficient for extensive methods' development;
Furthermore, such important facade classes as \textit{balcony} have no data representation.

Interestingly, the Oakland 3D \cite{Munoz-2009-10227} and the Paris-rue-Madame \cite{serna2014parisMadame} datasets also cover some aspects of the facade element classes.
Oakland 3D comprises only a few facade-related classes, inadequately capturing the key facade elements (e.g., windows are absent). 
Similarly, the Paris-rue-Madame dataset is restricted to merely wall lights, wall signs, and balcony plants, limiting its application for facade segmentation.
Another limitation of the datasets comprising facade classes is their limited size: 
TUM-FAÇADE and ArCH comprise around 100 million points, Paris-rue-Madame scores 20 million, and Oakland 3D has less than 2 million annotated points.  

Moreover, as shown in \Cref{tab:bigTable}, the class variability among the datasets is high, which hinders the standardized comparison between benchmarks and their reported segmentation results.
This trend is underscored by their high class-wise standard deviation of approximately 13, ranging from eight to up to 50 classes (excluding the non-labeled datasets). 
On the other hand, multiple international organizations exist that create standardized descriptions of urban objects, including facade elements.
One of the leading bodies is \gls*{OGC}, which has been releasing the international CityGML standard \cite{grogerOGCCityGeography2012,kolbe2008citygml}, providing a comprehensive description of geometries, structures, taxonomies, and aggregations at the scale of buildings but also entire countries \cite{Kolbe2021}.
Its wide adoption is highlighted by the example of more than 200 million open-data 3D building models available in Switzerland, Germany, the United States, the Netherlands, and Poland, among others \cite{wysocki2024reviewing}. 
The complementing sources for facade description are the Art and Architecture Thesaurus (AAT) and the \gls*{IFC} standard, which are widely applied in the architecture and civil engineering domains \cite{archDatasetPaper,laakso2012ifc}, for instance, in the building information modeling (BIM) context \cite{Kolbe2021}. 

The hierarchical understanding of the facade description has been also overlooked by all of the investigated datasets. 
Yet, in the context of building modeling standards three \gls*{LoDs} are recognized \cite{kutznerCityGMLNewFunctions2020}, acknowledging varying data availability, methods' performance on limited data, and the downstream task requirements.
Only Paris-Lille-3D \cite{roynard2018parisLille} follows the similar concept for point cloud segmentation, where they provide hierarchization of urban scenes. 
However, the classes do not describe facade elements, are not based on international standards, and many classes have none or limited data representation in the introduced data realization. 
\subsection{3D Facade Semantic Segmentation Methods}
The advent of the aforementioned image-based benchmarks has sparked advancements in facade segmentation using images.
Various methods have been proposed to tackle this challenge, starting with non-learning \cite{szeliski2010computer,musialski2013survey}, gramma-based \cite{MAYERejMCMC,brenner2006extraction}, and recently deep learning approaches \cite{liu2020deepfacade,KadaFacades,helmutMayerLoD3}.
However, 2D-image-reliance restricts the methods to 2D image information, which hampers its direct application to the 3D facade segmentation and thus limits capturing the facade intruded and extruded elements depth \cite{wysocki2023scan2lod3,KadaFacades}.
%

A different strategy focuses on direct 3D facade segmentation using laser scanning point clouds, leveraging the detailed and accurate depth information provided by \gls*{MLS} point clouds~\cite{xia2021vpc, xu2021towards}. 
Recently, the deep learning approaches have shown great potential in 3D point cloud segmentation~\cite{qi2017pointnet, qi2017pointnet++, zhao2021point}.
Qi et al., \cite{qi2017pointnet} present the PointNet architecture, which enables efficient point-wise 3D point cloud segmentation on unordered sets. 
Following works \cite{qi2017pointnet++,zhao2021point,wang2019dynamic}, have further improved segmentation performance.

Notably, significant advancements in point-wise, learning-based techniques have been applied to 3D facades segmentation as well~\cite{grilli2020machine, matrone2020comparing, yuetanDeepLearningOfficial, wysockiVisibility,BATTINI202437}. 
They typically perform well on the planar-like, ubiquitous classes, reaching around 75\% in F1 score for the wall class when applying DGCNN \cite{pierdicca2020point}.
However, the research shows the off-the-shelf methods face challenges when dealing with sparse- and under-represented classes, such as decorations, moldings, stairs, windows, and doors~\cite{matrone2020comparing,pierdicca2020point}.
Essentially, facing the typical long-tail data distribution recognition problem of real-world data \cite{zhang2023deep,flotzinger2024dacl,dai2017scannet,yeshwanth2023scannet++}.
This issue is reflected in, for example, low scores for the standard Point Transformer network \cite{zhao2021point}, which can merely reach F1 scores of 49\%, 2\%, and 48\%, for window, door, and molding class, respectively \cite{wysockiVisibility}.
Also, as reported by Pierdicca et al., recall scores for column, arc, decoration, door, and stair classes can oscillate around 0-1\% for PointNet and PointNet++\cite{pierdicca2020point}.

Yet, caution must be exercised while analyzing the reports, as the tests are currently inadequate, performed on heterogeneous classes and small datasets comprising a limited number of facades and their types \cite{tumfacadePaper,matrone2020comparing}; underscoring the need for developing classes harmonization and large-scale point cloud facade datasets.
The methods' performance is also dependent on the granularity of the target classes, where the reception field and sampling strategies play pivotal role \cite{thomas2019kpconv}.
This characteristic results in varying performance for dense-distribution objects (e.g., walls) and sparse and thin objects (e.g., windows) depending on network assumptions.
As such, the evaluation at various generalization facade levels is of pivotal importance.
%
\section{The LoFG and Its Realization}
\label{sec:tum-facade-pp}
\subsection{3D Semantic Facade Classes}
\begin{figure*}
    \centering
    \includegraphics[width=0.8\linewidth]{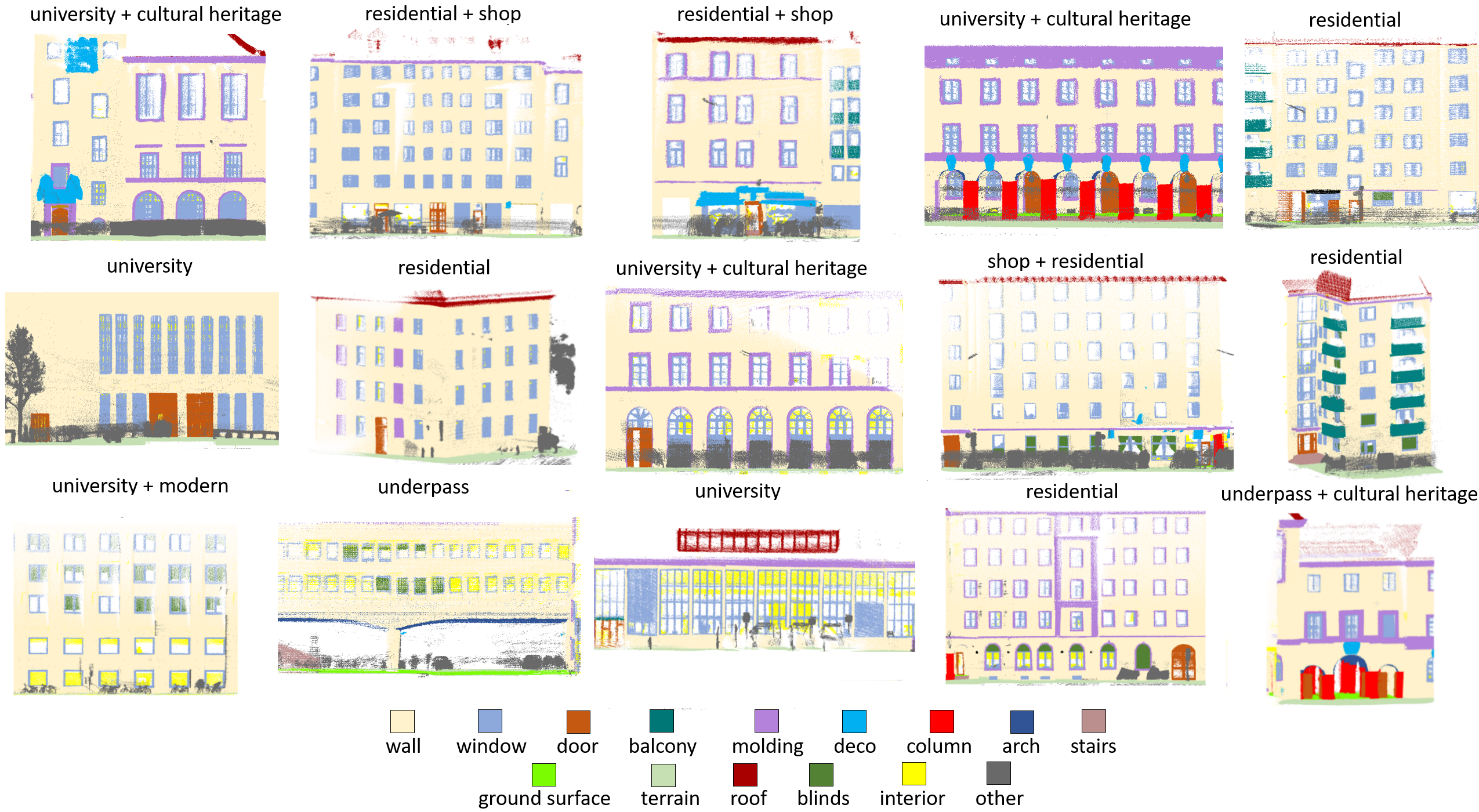}
    \caption{Selected facades of ZAHA. The dataset comprises 15 classes distributed over a diverse facade styles and functions, presenting a versatile training and testing segmentation scenario.}
    \label{fig:classes}
\end{figure*}
\begin{figure*}[h!t]
    \centering
    \includegraphics[width=0.8\linewidth]{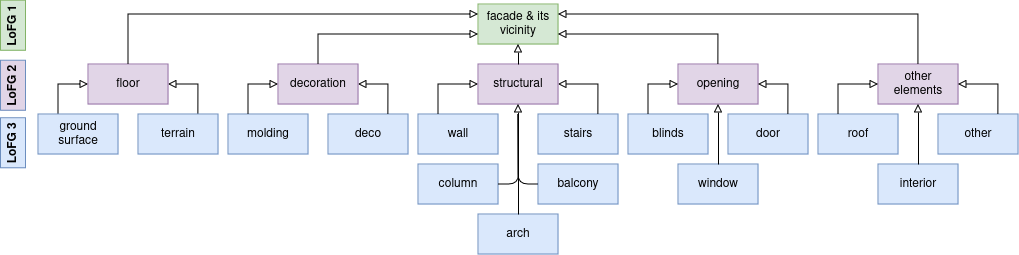}
    \caption{Level of Facade Generalization (LoFG): Primary 15 classes (blue) representing \gls*{LoFG} 3 at the finest level of generalization. These aggregate into coarser \gls*{LoFG} 2 (purple) based on their geometrical and semantical representation; The LoFG 1 (green) is designed as an abstract class allowing to group elements describing facade. Such hierarchical and harmonized representation offers adaptability to the downstream tasks of the built environment, addresses imbalanced facade datasets (long-tail), and allows for various but unified methods' testing.}
    \label{fig:genTree}
\end{figure*}

\begin{table*} 
    \captionsetup{size=footnotesize}
    \caption{Our proposed classes (\textit{Class}) for the ZAHA point cloud benchmark dataset. We aim to unify testing of 3D facade segmentation methods and directly link to the existing semantic 3D building modeling standards, such as CityGML \cite{grogerOGCCityGeography2012}.}
    \label{tab:identifiedClasses}
    \setlength\tabcolsep{5pt} 
    \footnotesize\centering
    \begin{adjustbox}{width=1\textwidth, center}
        \begin{tabular}{lll|ll}
            \toprule
            Index & Class & Description & Corresponding building & Corresponding building \\
            & & & standard class \cite{grogerOGCCityGeography2012} & standard function \cite{special_interest_group_3d_modeling_2020codelist,special_interest_group_3d_modeling_2020codelistEng} \\
            \midrule
            1 & wall & Excluding any decorative elements & WallSurface & - \\
            2 & window & Including on-surface reflections; excluding any decorative elements & Window & - \\
            3 & door & Including garage doors and gates & Door & - \\
            4 & balcony & Excluding columns and alike supportive structures & BuildingInstallation  & 1000, Balcony \\
            5 & molding & Decorative elements mainly adjacent to a facade (e.g., cornices) & BuildingInstallation & 1070, Other \\
            6 & deco & Decorative elements mounted to a facade (e.g., flags, gargoyles, lights) & BuildingInstallation  & 1070, Other \\
            7 & column & Excluding cornices (cornice → molding class) & BuildingInstallation & 1011, Column \\
            8 & arch & Including only surfaces oriented downwards (including outer ceilings) & BuildingInstallation & 1008, Arcade  \\
            9 & stairs & Excluding column alike supportive structures & BuildingInstallation & 1013, Stairs \\
            10 & ground surface & Any ground points inside a building envelope (e.g., thoroughfare) & GroundSurface & - \\
            11 & terrain & Any ground points outside a building envelope (e.g., sidewalks) & - & - \\
            12 & roof & Any surfaces representing a roof structure (incl. dormers) & RoofSurface & - \\
            13 & blinds & Window closures open or closed & BuildingInstallation & 1070, Other \\
            14 & interior & Measurements that reflect within a building envelope, excluding ground points & - & - \\
            15 & other & Any other elements, noise & - & - \\
            \bottomrule
        \end{tabular}
    \end{adjustbox}
\end{table*}
As shown in \Cref{tab:bigTable}, there is a lack of high facade variability and hierarchical facade-level point cloud benchmarks available.
Understanding the need for 3D facade semantic segmentation gradation due to different methods' assumptions, downstream tasks, and data availability, we introduce \gls*{LoFG}, where we distinguish three levels of facade generalization: \gls*{LoFG}1, \gls*{LoFG}2, and \gls*{LoFG}3.
This concept and its presented logic in \cref{fig:genTree} aim to allow for precisely formulating 3D facade semantic segmentation and classification problems.
Furthermore, it shall enable seamless adoption of such techniques as transfer learning \cite{shen2023segtrans}, epoch-to-epoch labels transferring \cite{schwarz2023transferring}, and a unified comparison of different methods.
%

Within the scope of this work, we introduce the harmonized 15 facade classes (\gls*{LoFG}3) generalizing to five classes (\gls*{LoFG}2), and one abstract class (\gls*{LoFG}1), developed concerning the international urban modeling standards such as CityGML, \gls*{IFC}, and Art and Architecture Thesaurus (AAT), and related works \cite{tumfacadePaper,matrone2020comparing}.
The detailed description of the introduced classes is provided in \cref{tab:identifiedClasses} whereas the visualization is in \cref{fig:catchy} and \cref{fig:classes}.
Moreover, owing to leveraging the modeling standards the presented classes correspond to the established 3D semantic reconstruction classes, aiming to provide a seamless platform for applying the 3D facade semantic segmentation results to the 3D facade semantic reconstruction \cite{grogerOGCCityGeography2012, kutznerCityGMLNewFunctions2020,wysocki2023scan2lod3}. 

The \gls*{LoFG}3 describes the most detailed facade representation comprising 15 facade classes of \textit{ground surface}, \textit{terrain}, \textit{molding}, \textit{deco}, \textit{wall}, \textit{stairs}, \textit{balcony}, \textit{column}, \textit{arch}, \textit{blinds}, \textit{door}, \textit{window}, \textit{roof}, \textit{interior}, and \textit{other} (for the detailed description see \cref{tab:identifiedClasses}). 
The \gls*{LoFG}2 aggregates the 15 classes of \gls*{LoFG}3 into five less detailed classes based on syntactic, semantic, and geometrical analysis, as we illustrate in \cref{fig:genTree}. 
The group \textit{facade \& its vicinity} represents the abstract class and is referred to as \gls*{LoFG}1.

Note that the \gls*{LoFG} is driven mainly by the semantics of facade elements and their structural definition and not sole geometrical representation.
For example, \textit{roof} is often regarded as a structural element of a building; however, in the case of terrestrial-acquired datasets, roofs are barely within the scanner field-of-view and thus resemble noise, for example, shown in \cite{tumfacadePaper} or in \cref{fig:catchy}, \cref{fig:classes}.
Furthermore, the ground and terrain classes are also clearly defined in the international standards and are considered key parts of a facade, e.g., the intersection of \textit{terrain} with a facade is essential in establishing the total facade height, while \textit{ground surface} is vital in analyzing overarching structures and their volumetric extent, e.g., underpasses.
Another example are the classes \textit{windows} and \textit{blinds}, which at \gls*{LoFG}3 are separate, owing to they different features, functions, and importance of such separation for window segmentation \cite{tuttas_reconstruction_2013,wysocki2023scan2lod3};
Yet, at \gls*{LoFG}2 they are aggregated, as a blind is indissolubly linked to a window.

%
%
\subsection{Data Acquisition}
Based on the conducted research (\cref{tab:bigTable}) and seeing the potential of already created urban-related point cloud benchmarks, we present an approach to reducing the workload while developing new benchmark datasets. 
This reduction is achieved by enriching existing benchmarks with facade-related semantics.
The data acquired for the ZAHA dataset stems from the open dataset of TUM-MLS-2016\cite{zhu_tum-mls-2016_2020}, featuring a challenging, urban environment with real-world, dense, and georeferenced \gls*{MLS} point clouds.
The measuring campaign is performed within the city of Munich, Germany.
We utilize this dataset owing to its versatile architectural style of buildings built between the late 19th and early 21st century and different facade functions ranging from regular dwellings, educational, cultural heritage, shops, and traffic underpasses.
Notably, the datasets are subject to active development, including \gls*{LoD}3 building models, which may be used as an additional validation set \cite{tum2twin}.
\subsubsection{Mobile Laser Scans}
The used TUM-MLS-2016 relies on the Mobile Distributed Situation Awareness (MODISSA) platform, employing two Velodyne HDL-64E LiDAR sensors obliquely mounted at the front and two Velodyne VLP-16 sensors at the rear of the van-type vehicle. 
The inertial navigation system is complemented by real-time kinematic (RTK) correction data from the German satellite positioning service (SAPOS), enhancing accurate georeferencing throughout the data collection process \cite{borgmann2018data, zhu_tum-mls-2016_2020}.
Due to large float numbers, we present the data in a local \gls*{CRS} and attach the transformation matrix, which enables back-transformation of the projected \gls*{CRS} of UTM 32 (EPSG: 25832).
\subsubsection{Semantic Annotation}
We leverage the georeferencing to support the annotation and data extraction process.
We acquire cm-grade footprints from governmental, open-data CityGML \gls*{LoD}2 building models \cite{tumLoD2bavariaGov}, which are available in the same \gls*{CRS} as the obtained point clouds.
To extract the facades and their vicinity, we draw a buffer around each building footprint with a radius of 3m.
Furthermore, in this process, each point cloud obtains its corresponding unique global identification (ID) of the building entity, matching the governmental database and thus enabling model-to-point-cloud comparison. 

The manual annotation of point cloud entities is performed using Semantic Segmentation Editor \cite{hitachiEditor}.
The point clouds are divided into batches of approximately four million points, considering software and hardware capabilities and the operator's ability to discern various facade features.
We expand the software annotation set to accommodate our specific classes, outlined in \cref{tab:identifiedClasses}, and create the setup file available in our repository.
After the point clouds are merged back into entities, another round of manual inspection is applied to minimize the manually induced errors between batches.
Here, we employ another software, Cyclone 3DR \cite{cyclone}, to reduce errors induced by differences in visualization and annotation tools.
For the first round of annotations, we estimate that depending on the complexity of an object, labeling requires between seven to 23 hours per building, averaging approximately six hours per facade.
The following correction round of annotations requires approximately two hours per facade.
\subsection{Main Benchmark Challenges}
We believe that the main challenges while developing the methods performing using the ZAHA classes and dataset are as follows:
\begin{itemize}
    \item \textbf{Classes relevant to the built environment} We introduce \gls*{LoFG} that harmonizes the so-far unstructured facade element classes, whereby we utilize the facade-related international standards. This characteristic ensures a homogeneous comparison of the developed algorithms and exposes the dataset to the practical challenges of the built environment. Furthermore, different levels of generalization allow the testing of various methods' assumptions. Such design also challenges the current methods, as such classes are often highly imbalanced, see the common long-tail recognition issue in \cite{zhang2023deep,yeshwanth2023scannet++,dai2017scannet} and in our \cref{fig:longTail}.  
\begin{figure}[h!t]
    \centering
    \includegraphics[width=\linewidth]{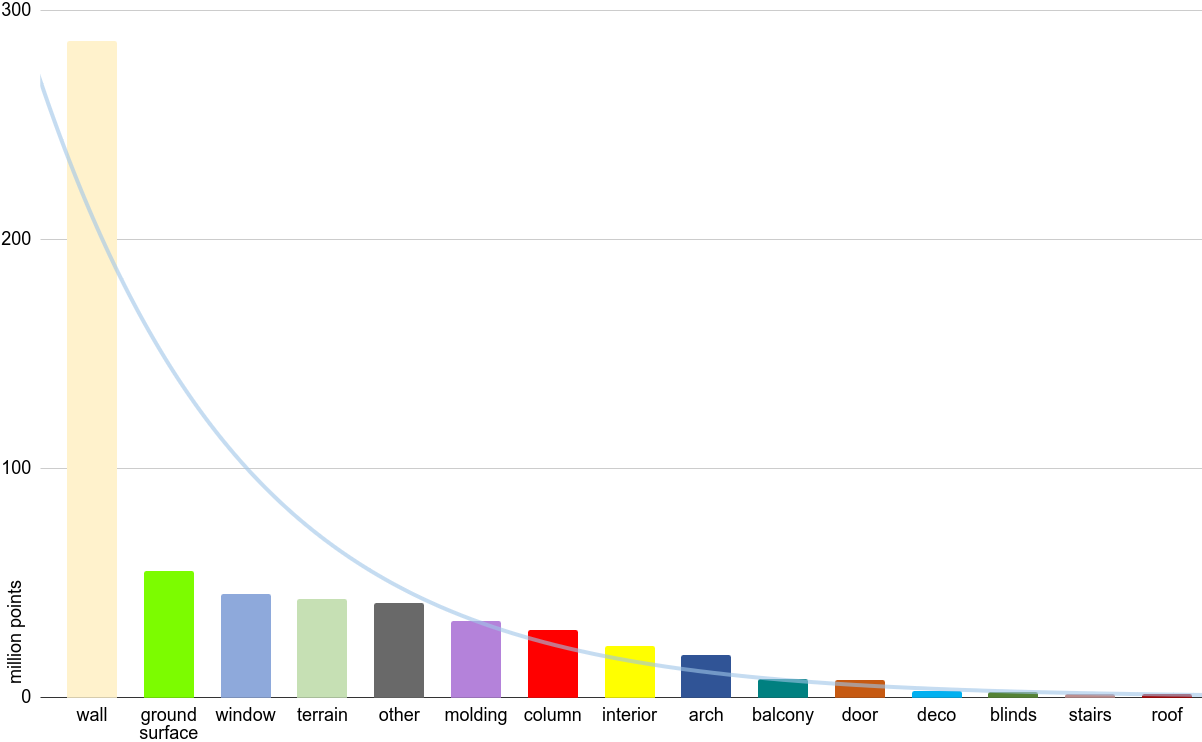}
    \caption{Head and long-tail (blue line) point-per-class distribution in ZAHA: The typical challenge of the real-world recognition tasks.}
    \label{fig:longTail}
\end{figure}
    \item \textbf{Facade type variations} Unlike the other facade-related datasets, we present 66 facades in various architectural styles and function types, having more than four times as many points for training and validation when compared to the largest datasets to date \cite{archDatasetPaper,tumfacadePaper}.
    This trait allows testing the generalization capabilities of 3D facade semantic segmentation algorithms. Simultaneously, it poses a scientific challenge in designing a generic method agnostic to the architectural facade types.
    \item \textbf{Consumer-grade \gls*{MLS} measurements} We provide the non-filtered point cloud acquired with an off-the-shelf LiDAR devices (Velodyne). No special noise corrections and dynamic object removal are applied; interior reflections and adjacent to facade objects are kept, too. Furthermore, the given point cloud comprises only geometrical representation and no spectral information. As such, the dataset yields a challenging, real-world, and raw point clouds, useful for testing not only facade but also any generic semantic segmentation method. 
\end{itemize}
%
\section{Experiments}
We conducted the experiments on the ZAHA benchmark dataset comprising 601~million semantic-annotated points%
\footnote{Project page: https://github.com/OloOcki/zaha}. 
The validation set comprised typical residential facades, educational and cultural heritage facades, and facades with an underpass. 
The test set reflected in functions the validation set.
We designated the rest of the facades for training.
In our training setup, we also ensured that training, validation, and test data subsets each covered all 15 (\gls*{LoFG}3) and five (\gls*{LoFG}2) introduced classes; 
Consequently, the applied training and testing methods were exposed to each facade class.
We also evaluated the introduced \gls*{LoFG} classes by applying the same training routine to both \gls*{LoFG}2 and \gls*{LoFG}3 levels and comparing the results.
%
%





%
%
\subsection{Baseline Semantic Segmentation Methods}
To investigate the unresolved 3D facade segmentation challenges, we tested a set of well-established semantic segmentation networks and metrics along with their original implementations on ZAHA (see the supplement for more settings' details).
To evaluate the networks' performance, we used Overall Accuracy, Precision, Recall, and Jaccard Index, also known as Intersection over Union (IoU) \cite{li2020deep,wysockiVisibility,pierdicca2020point}.
Our methods selection was not only dictated by their performance reported in other urban-related works and datasets \cite{wang2023building3d,tan2020toronto,matrone2020comparing,pierdicca2020point,grilli2020machine} but also by their feature extraction principles:
%
%
%
\begin{itemize}
    \item \textbf{PoinNet} presented by Qi et al., \cite{qi2017pointnet} is a seminal work in a direct point cloud understanding. The method is permutation invariant, learns point-wise local features, and allows for understanding the global scenes.  
    \item \textbf{PointNet++}\cite{qi2017pointnet++} improves its PoinNet predecessor by hierarchical understanding. Thus enabling more effective capture of local structures, especially in the context of fine-grained object segmentation. 
    \item \textbf{Point Transformer (PT)}  \cite{zhao2021point} utilizes the language-understanding-inspired self-attention mechanism, weighing the importance of different elements in a sequence when processing each element. It also has positional encoding, as the transformers do not capture the point position but rather sequences.
    \item \textbf{DGCNN} \cite{wang2019dynamic} leverages graph structures for learning the point set semantics, as it constructs dynamic edge convolutional graphs based on the input point clouds. It uses the k-nearest neighbors graph to capture local structures and a farthest point sampling strategy to aggregate global context.  
\end{itemize}
\section{Results and Discussion}
%
\begin{figure}[ht]
    \centering
    \includegraphics[width=\linewidth]{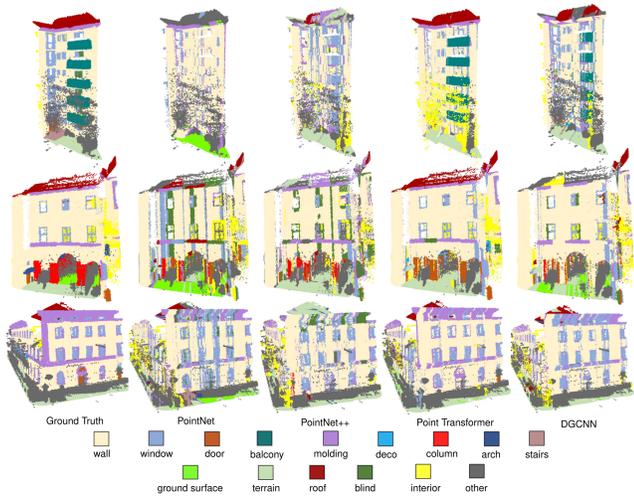}
    \caption{LoFG3 inference on the test set comprising residential (top), underpass and cultural heritage (middle), and university buildings (bottom).  }
    \label{fig:inferenceLoFG3}
\end{figure}
\begin{table}[h]
    \setlength\tabcolsep{3.5pt} 
    \scriptsize
    \centering
    \caption{LoFG3 test: OA, $\mu$P, $\mu$R, $\mu$F1, $\mu$IoU, and F1 scores per class (in percentages, top scores bold).}
    \label{tab:testLoFG3}
    \begin{tabular}{l|ccccccc}
        \hline
        \textbf{Method} & PointNet & PointNet++ & Point Transformer & DGCNN \\
        \hline
        OA & 59.9 & 66.4 & \textbf{75.0} & 71.1 \\
        $\mu$P & 46.1 & 37.8 & 52.7 & \textbf{53.6} \\
        $\mu$R & 42.2 & 35.9 & \textbf{54.7} & 45.8 \\
        $\mu$F1 & 38.7 & 34.8 & \textbf{52.1} & 44.5 \\
        $\mu$IoU & 26.4 & 25.6 & \textbf{41.6} & 33.4 \\ \hline
        wall & 61.1 & 68.5 & 76.8 & \textbf{83.8} \\
        window & 25.6 & 26.3 & 43.1 & \textbf{64.1} \\
        door & 13.5 & 7.8 & 19.8 & \textbf{21.6} \\
        balcony & 25.1 & 0.0 & \textbf{77.5} & 66.7 \\
        molding & 22.5 & 43.4 & \textbf{58.0} & 57.5 \\
        deco & 0.0 & 0.0 & \textbf{5.0} & 0.0 \\
        column & 22.4 & 33.4 & 0.0 & \textbf{37.2} \\
        arch & 19.2 & 25.4 & \textbf{50.2} & 2.6 \\
        stairs & \textbf{16.0} & 0.0 & 7.5 & 5.6 \\
        ground surface & 12.0 & 0.0 & \textbf{24.4} & 21.3 \\
        terrain & 53.5 & 53.5 & 57.6 &  \textbf{68.0} \\
        roof & 18.7 & 6.8 & \textbf{66.3} & 57.4 \\
        blinds & 4.6 & 2.3 & 18.5 & \textbf{20.0} \\       
        interior & 59.7 & 69.1 & 72.8 & \textbf{88.0} \\
        other & 42.7 & 47.1 & 70.6 & \textbf{74.1} \\
        \hline
    \end{tabular}
\end{table}
%
\subsection{Well-Performing Class Segmentation} 
\begin{figure}
    \centering
    \includegraphics[width=\linewidth]{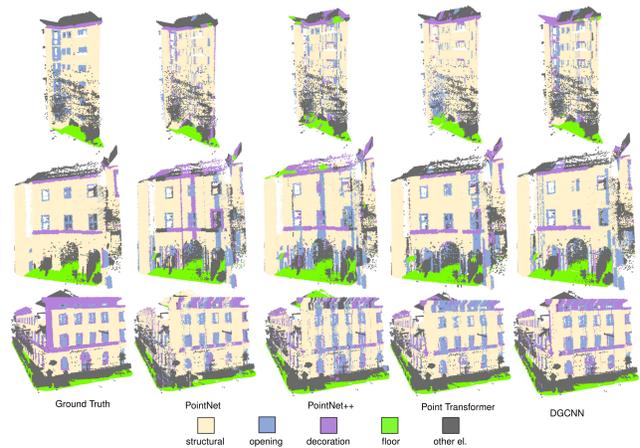}
    \caption{LoFG2 inference on the test set comprising residential (top), underpass and cultural heritage (middle), and university buildings (bottom); color-coding according to the most prominent merged sub-class.}
    \label{fig:inferenceLoFG2}
\end{figure}

Our experiments corroborate the research consensus that classes that are well-represented and characterized by planar geometries are likely to be correctly segmented.
As we show in \Cref{fig:inferenceLoFG3} and \Cref{fig:inferenceLoFG2}, this trend is reflected in both generalization levels \gls*{LoFG}2 and \gls*{LoFG}3.
The most prominent example for the \gls*{LoFG}3 is the \textit{wall} class, which was the overall best-performing class with a median score of 73\% (\cref{tab:testLoFG3}) across all the applied networks.
For the \gls*{LoFG}2, it was the \textit{floor} class comprising primarily planar-like objects that achieved a high median F1 score, reaching approximately 91\% across the tested baseline methods (\cref{tab:testLoFG2}).
Interestingly, also \textit{interior} and \textit{other} classes were reliably distinguished with up to 88\% and 74\% F1 scores, respectively (\cref{tab:testLoFG3}).
They differentiate themselves strongly by exposing highly unstructured local patterns followed by structured global patterns, i.e., outside-inside of a facade. 
\subsection{Challenging Class Segmentation} Confirming our hypothesis, one of the most challenging classes was \textit{deco}, scoring at best approximately 5\% (\cref{tab:testLoFG3}) and remained largely unsegmented, as shown in \cref{fig:inferenceLoFG3}.
We attribute it to the high structure complexity, uniqueness among the samples, and high under-representation, i.e., being at the long tail of sample distribution (see \cref{fig:longTail} for the distribution).
When aggregated in \gls*{LoFG}2 with \textit{molding}, the score increased and reached, on average, 48\%; as such, however, remaining an unresolved challenge.

As shown in \cref{fig:inferenceLoFG3} and \cref{fig:inferenceLoFG2}, the \textit{window} and \textit{door} classes, which are the most prominent facade features, yet typically label-sparse due to their translucent main parts, were merely partially segmented.
As our experiments in \cref{tab:testLoFG3} show, \textit{window} at best scored 64\%, while \textit{door} merely 22\%;
Also, the related \textit{blinds} class scored at best 20\%.
When generalized to \gls*{LoFG}2 as \textit{opening}, their performance reached at best 66\%.
Oscillating in the range of scores of 50\% and less were also \textit{arch}, \textit{stairs}, \textit{ground surface}, \textit{column}, \textit{molding}, and \textit{roof}; underlining the need for designing more robust 3D facade segmentation methods.
\subsection{Level of Facade Generalization (LoFG)}
Our experiments show the usability of the proposed aggregation hierarchy design as a concept addressing 3D semantic facade segmentation at different generalization levels while maintaining facade-related semantics.
\begin{table}[h]
    \setlength\tabcolsep{3.5pt} 
    \scriptsize
    \centering
    \caption{LoFG2 test: OA, $\mu$P, $\mu$R, $\mu$F1, $\mu$IoU, and F1 scores per class(in percentages, top scores bold).}
    \begin{tabular}{l|ccccc}
        \hline
        \textbf{Method} & PointNet & PointNet++ & Point Transformer & DGCNN \\
        \hline
        OA & 71.9 & 75.5 & 78.2 & \textbf{82.6} \\
        $\mu$P & 69.6 & 73.0 & 75.8 & \textbf{80.0} \\
        $\mu$R & 68.1 & 73.0 & 76.6 & \textbf{81.8} \\
        $\mu$F1 & 68.1 & 72.6 & 76.1 & \textbf{80.4} \\
        $\mu$IoU & 55.8 & 59.8 & 63.9 & \textbf{68.5} \\ \hline
        floor & \textbf{92.3} & 87.6 & 90.7 & 92.1 \\
        decoration & 26.2 & 47.1 & 47.0 & \textbf{70.0} \\
        structural & 60.9 & 65.5 & 67.0 & \textbf{85.2} \\
        opening & 28.2 & 27.2 & 36.0 & \textbf{66.2} \\
        other el. & 71.2 & 71.6 & 78.9 & \textbf{88.8} \\
        \hline
    \end{tabular}
    \label{tab:testLoFG2}
\end{table}
As expected, all the metric scores increased when testing on aggregated \gls*{LoFG}2 instead of high-detail \gls*{LoFG}3, with OA difference maximum for PointNet of 12\% and the median of approximately 10\% across the tested methods (\cref{tab:testLoFG2} and \cref{tab:testLoFG3}).
When analyzing \gls*{LoFG}2 only, we deem the \textit{decoration} and \textit{opening} the most challenging classes owing to their low-performance median scores of 47\% and 32\%, respectively.
On the other hand, we observe high to medium performance for \textit{floor}, \textit{structural}, and \textit{other elements} classes, that median scores were approximately 91\%, 66\%, 75\%, respectively. 

Notably networks performance ranking also varied depending on the generalization level, corroborating our assumption that methods' performance is largely dependent on the segmentation objective classes:
At \gls*{LoFG}2 the DGCNN outperformed the Point Transformer network (e.g., in recall by around 5\%), and vice versa for \gls*{LoFG}3 (e.g., in recall by around 9\%).
Per-class analysis at \gls*{LoFG}3 also unveils that transformer-based network (PT) can have higher scores on different classes than the graph-based network (DGCNN).
%
%
\section{Conclusion}
In this work, we present ZAHA: a) The hierarchical segmentation classes for facade segmentation, called Level of Facade Generalization (LoFG), designed based on the international facade-related standards;
b) Complemented by the classes realization on the to date largest, real-world, large-scale point cloud facade benchmark data comprising approximately 601 million points, surpassing the current largest facade benchmark four times.

The findings of this study indicate that 3D facade segmentation remains challenging and necessitates comprehensive benchmark data and unified classes introduced by LoFG.
The semantic segmentation methods perform well on planar-like facade elements (up to 84\% for \textit{wall}) but struggle on intricate and sparsely represented objects (up to 5\% for \textit{deco}).
We also observe that the critical elements of a facade, such as a door and a window, still necessitate novel segmentation methods (accuracy up to 22\% for the door, 64\% for the window).
Moreover, the performance of the neural networks largely varies depending on the selected generalization level: At \gls*{LoFG}2 the graph-based DGCNN outperformed transformer-based Point Transformer, and inversely for \gls*{LoFG}3; underscoring the need for evaluation at different generalization levels.


Based on the observed trend in the 2D image-based facade segmentation domain, we firmly believe this 3D facade segmentation dataset will foster further development of 3D facade-oriented methods.
Consequently, unlocking various downstream and related tasks, such as robust 3D semantic facade reconstruction for autonomous driving testing \cite{schwabRequirementAnalysis3d2019} or flood damage assessment \cite{nouvel2013citygml}.
We plan to extend our work by organizing a 3D facade semantic segmentation challenge and leaderboard\footnote{Leaderboard: https://tum2t.win/benchmarks/pc-fac} to further facilitate these developments.

{\bf Acknowledgments}
The work was conducted within the framework of the Leonhard Obermeyer Center at the Technical University of Munich (TUM).
The work on this project commenced already in 2020 and was accomplished thanks to support of various groups and researchers:
The special mention goes to Jiarui Zhang, Yue Tan, Chenkun Zhang, and Prabin Gyawali, for their diligent work in the annotation process.

{\small
\bibliographystyle{ieee_fullname}
\bibliography{main}
}

\end{document}


\title{The supplementary material to: \\ "ZAHA: Introducing the Level of Facade Generalization and the Large-Scale Point Cloud Facade Semantic Segmentation Benchmark Dataset"}  

\maketitle
\thispagestyle{empty}
\appendix

\section{Experiments}
\label{sec:intro}
\sloppy

\subsection{Evaluation Metrics}

To evaluate the performance of the 3D facade segmentation, we used the established semantic segmentation network metrics, such as Overall Accuraccy, Precision, Recall, and Jaccard Index also known as Intersection over Union (IoU) \cite{li2020deep}.
They were defined as follows:

\[
\text{Overall Accuracy} = \frac{\text{True Positives + True Negatives}}{\text{Total Instances}}
\]

\[
\text{Precision} = \frac{\text{True Positives}}{\text{True Positives + False Positives}}
\]

\[
\text{Recall} = \frac{\text{True Positives}}{\text{True Positives + False Negatives}}
\]

\[
F1 = 2 \times \frac{\text{Precision} \times \text{Recall}}{\text{Precision + Recall}}
\]

\[
\text{IoU} = {\frac{\text{Intersection Area}}{\text{Union Area}} = }
\]
\[
 = \frac{\text{True Positives}}{\text{True Positives + False Positives + False Negatives}}
 \]
\subsection{Parameter Settings}
We conducted all the experiments using an NVIDIA GeForce RTX 4090 GPU with 16 GB VRAM with a fixed number of 100 epochs per training. 
The implementation will be released under our repository web page.%
\footnote{Project page: https://github.com/OloOcki/zaha}

To train the PointNet and PointNet++, we utilized the implementation sourced from \cite{qi2017pointnet, Pytorch_Pointnet_Pointnet2, qi2017pointnet++}.
We used the point cloud coordinates as input layers to adapt the model and modify the corresponding classes.
We employed a batch size of 32 for training and testing, and each batch contains 1024 points per sample point cloud.
Stochastic gradient descent with a momentum of 0.1 and a learning rate 0.001 was employed.

To train the Point Transformer, we used the implementation of the first author of Point Transformer\cite{zhao2021point}, in \cite{ptingithub}. 
Diverging from the voxelization method employed in the source code to generate batches, we have opted for a batch design strategy based on index segmentation.
Consistent with previous experiments, we also utilized a batch size of 32 as input and 1024 points per sample point cloud.
Stochastic gradient descent with a momentum of 0.9 and an initial learning rate of 0.1 was employed. 
After 60 epochs, the learning rate would reduced to 0.01, and after 80 epochs, 0.001.

To train the DGCNN, we utilized the implementation sourced from \cite{wang2019dynamic,dgcnnPytorch}.
We customized their implementation by adjusting the dimensions of the input and output layers to suit the dimensionality and the number of classes.
During training, we employed a batch size of 32, while for testing, we used a batch size of 16, each containing 1024 points per sample point cloud.
Stochastic gradient descent with a momentum of 0.9 and a learning rate of 0.1 was employed.
We set the dropout rate to 0.5 and the number of nearest neighbors that we considered to 20.
\subsection{Extra Baseline Experiment on a Large-Scale-Oriented Network}
\begin{figure*}[ht]
    \centering
    \includegraphics[width=0.8\linewidth]{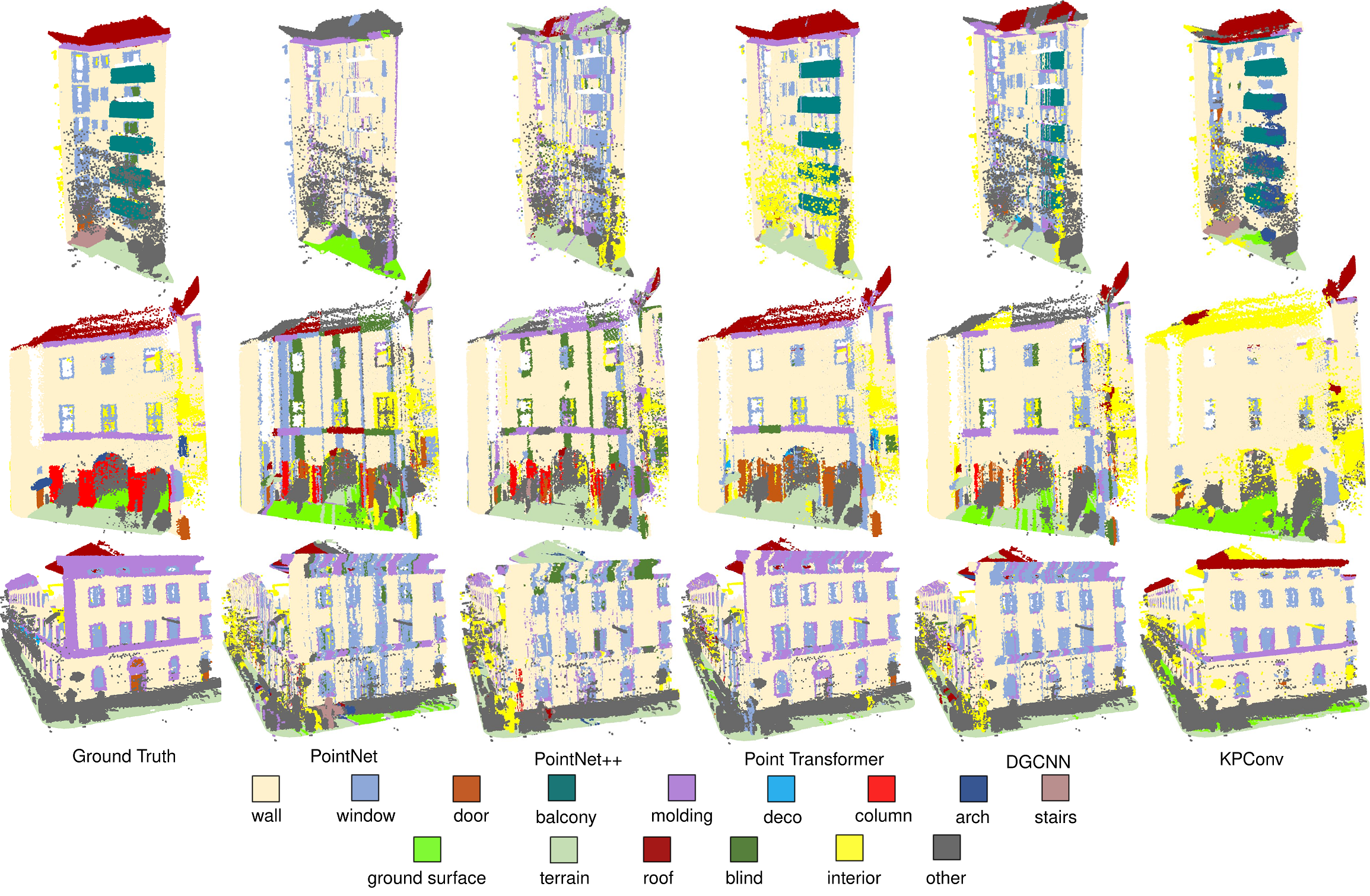}
    \caption{LoFG3 inference on the test set comprising residential (top), underpass and cultural heritage (middle), and university buildings (bottom) with the extra fine-tuned KPConv.  }
    \label{fig:inferenceLoFG3}
\end{figure*}
\begin{figure*}
    \centering
    \includegraphics[width=0.8\linewidth]{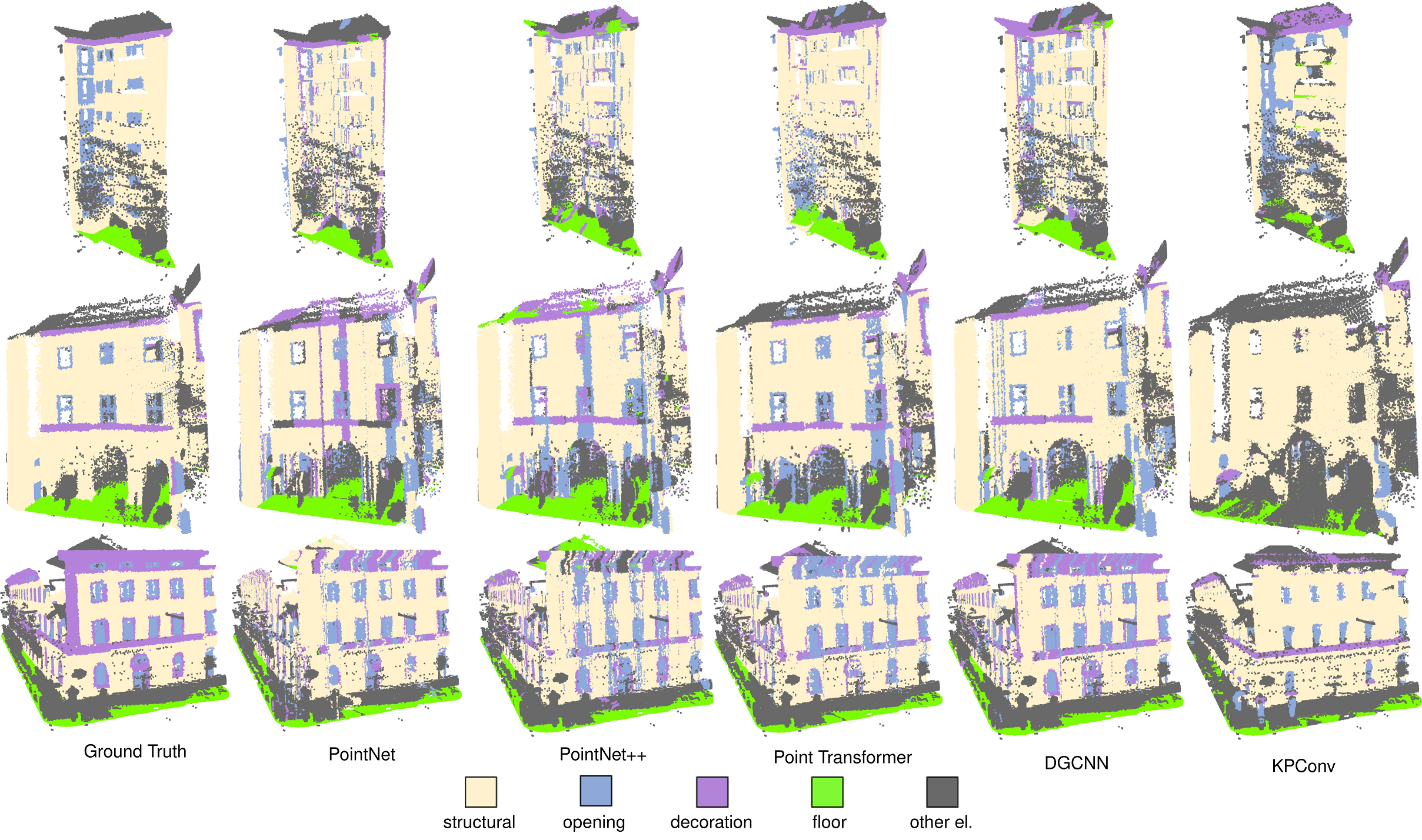}
    \caption{LoFG2 inference on the test set comprising residential (top), underpass and cultural heritage (middle), and university buildings (bottom); color-coding according to the most prominent merged sub-class with the extra fine-tuned KPConv.}
    \label{fig:inferenceLoFG2}
\end{figure*}
Owing to the space limitation and similar performance scores to the other networks, we have moved the extra experiments on the large-scale-oriented KPConv \cite{thomas2019kpconv} network to the supplemental material. 
Here, we show the extra set of experiments that we conducted on the KPConv network, whereby we also fine-tuned the hyper-parameters. 
KPConv introduces a deformable convolution operation, allowing the neural network to learn flexible and adaptive convolutional filters.
\begin{table}[h]
    \setlength\tabcolsep{3.5pt} 
    \scriptsize
    \centering
    \caption{LoFG3 test: OA, $\mu$P, $\mu$R, $\mu$F1, $\mu$IoU, and F1 scores per class, in percentages; with the extra fine-tuned KPConv}
    \label{tab:testLoFG3}
    \begin{tabular}{l|cccccccccccccccccccccccc}
        \hline
        \textbf{Method} & PointNet & PointNet++ & Point Transformer & DGCNN & KPConv \\
        \hline
        OA & 59.9 & 66.4 & 75.0 & 71.1 & 65.2 \\
        $\mu$P & 46.1 & 37.8 & 52.7 & 53.6 & 46.4 \\
        $\mu$R & 42.2 & 35.9 & 54.7 & 45.8 & 44.6 \\
        $\mu$F1 & 38.7 & 34.8 & 52.1 & 44.5 & 39.3 \\
        $\mu$IoU & 26.4 & 25.6 & 41.6 & 33.4 & 28.6 \\ \hline
        wall & 61.1 & 68.5 & 76.8 & 83.8 & 66.6 \\
        window & 25.6 & 26.3 & 43.1 & 64.1 & 41.0 \\
        door & 13.5 & 7.8 & 19.8 & 21.6 & 9.6 \\
        balcony & 25.1 & 0.0 & 77.5 & 66.7 & 61.7 \\
        molding & 22.5 & 43.4 & 58.0 & 57.5 & 23.3 \\
        deco & 0.0 & 0.0 & 5.0 & 0.0 & 0.0 \\
        column & 22.4 & 33.4 & 0.0 & 37.2 & 0.0  \\
        arch & 19.2 & 25.4 & 50.2 & 2.6 & 11.5 \\
        stairs & 16.0 & 0.0 & 7.5 & 5.6 & 6.5 \\
        ground surface & 12.0 & 0.0 & 24.4 & 21.3 & 26.3 \\
        terrain & 53.5 & 53.5 & 57.6 & 68.0 & 31.4 \\
        roof & 18.7 & 6.8 & 66.3 & 57.4 & 15.4 \\
        blinds & 4.6 & 2.3 & 18.5 & 20.0 & 10.0 \\       
        interior & 59.7 & 69.1 & 72.8 & 88.0 & 75.1 \\
        other & 42.7 & 47.1 & 70.6 & 74.1 &  50.0 \\
        \hline
    \end{tabular}
\end{table}
\begin{table}[h]
    \setlength\tabcolsep{3.5pt} 
    \scriptsize
    \centering
    \caption{LoFG2 test: OA, $\mu$P, $\mu$R, $\mu$F1, $\mu$IoU, and F1 scores per class, in percentages; with the extra fine-tuned KPConv}
    \begin{tabular}{l|cccccc}
        \hline
        \textbf{Method} & PointNet & PointNet++ & Point Transformer & DGCNN &  KPConv\\
        \hline
        OA & 71.9 & 75.5 & 78.2 & 82.6 & 71.6\\
        $\mu$P & 69.6 & 73.0 & 75.8 &  80.0 & 71.2 \\
        $\mu$R & 68.1 & 73.0 & 76.6 & 81.8 & 64.3 \\
        $\mu$F1 & 68.1 & 72.6 & 76.1 &  80.4 & 66.4 \\
        $\mu$IoU & 55.8 & 59.8 & 63.9 &  68.5 & 52.3 \\ \hline
        floor & 92.3 & 87.6 & 90.7 &  92.1 & 80.1 \\
        decoration & 26.2 & 47.1 & 47.0 & 70.0 & 28.2 \\
        structural & 60.9 & 65.5 & 67.0 &  85.2 & 62.4 \\
        opening & 28.2 & 27.2 & 36.0 &  66.2 & 31.7 \\
        other el. & 71.2 & 71.6 & 78.9 &  88.8 & 58.5 \\
        \hline
    \end{tabular}
    \label{tab:testLoFG2}
\end{table}
The use of kernel points in KPConv allows for more efficient processing of point clouds, and as such, it has often been used in the context of large-scale, outdoor point clouds \cite{kada2021point}.
However, corroborating our experiment results in the main paper, there were no significant performance differences observed, as we visualize in \Cref{fig:inferenceLoFG3} and \Cref{fig:inferenceLoFG2}, and list in \Cref{tab:testLoFG3} and \Cref{tab:testLoFG2}.  

To train the KPConv, we employed the implementation sourced from \cite{thomas2019kpconv,KPConvPytorch}.
The input radius of the input sphere was set as 1.5m. 
We also generated the radius of deformable convolution in the "number grid cell" as 1.5m, to minimize noisy clusters. 
For both baselines, we set the epoch steps as 2000 and the validation size as 100. 
The batch size was set to 6 for training and 1 for testing.
For the training of LoFG2, stochastic gradient descent with a momentum of 0.98 and a learning rate of 0.001 was employed, and there were 150 training epochs.
For the training of LoFG3, stochastic gradient descent with a momentum of 0.98 and a learning rate of 0.01 was employed, and the training epochs were set as the default number 500.

\subsection{Benchmark and Leaderboard}
We introduce the ZAHA as a benchmark to foster the research on facade semantic segmentation. 
It is a common practice to publish a leaderboard, which encourages researchers to delve into a challenge. 
We initialize the leaderboard at the webpage \footnote{Leaderboard: https://tum2t.win/benchmarks/pc-fac} and invite researchers to develop novel and more efficient facade segmentation methods.

\subsection{Extra visuals}
Additional visuals are included at the project page \footnote{Project page: https://github.com/OloOcki/zaha} showing an animated gif file, and the full dataset and annotations according to the 15 introduced classes at LoFG3 and their generalization at LoFG2.

{\small
\bibliographystyle{ieee_fullname}
\bibliography{supplement}
}